\documentclass[journal,onecolumn]{IEEEtran}
\ifCLASSINFOpdf
\else
\fi

\usepackage{booktabs}
\usepackage{multirow}
\usepackage{amsmath}

\usepackage{graphicx}

\begin{document}
\title{A Reinforcement learning method for Optical Thin-Film Design}

\author{Anqing Jiang,
        Liangyao Chen,
        Osamu Yoshie}% <-this % stops a space}

\maketitle

\begin{abstract}
Machine learning, especially deep learning, is dramatically changing the methods associated with optical thin-film inverse design. The vast majority of this research has focused on the parameter optimization (layer thickness, and structure size) of optical thin-films. A challenging problem that arises is an automated material search. In this work, we propose a new end-to-end algorithm for optical thin-film inverse design. This method combines the ability of unsupervised learning, reinforcement learning(RL) and includes a genetic algorithm to design an optical thin-film without any human intervention. Furthermore, with several concrete examples, we have shown how one can use this technique to optimize the spectra of a multi-layer solar absorber device. 
\end{abstract}


% \fi
\IEEEpeerreviewmaketitle

\section{Introduction}
In recent decades, significant fundamental advances combined with the spectacular progress of nano-scale fabrication methods have led to a broad range of innovations in the design of optical thin-film. Many applications, such as broadband filter\cite{yang2016compact, li2019ultra, ma2019high}, solar absorber\cite{gao2019structure, khoza2019structural, rubin2019optical}, and radiative cooling device\cite{chae2020spectrally, naghshine2018optimized}, increasingly rely on the intricate nanostructure design for greater performance at target wavelengths. Most of researchers make such designs based on human intelligence to solve a fundamental photonic problem: choosing the best combination of materials and nano-structure of a layered optical thin-film. Human intelligence is often limited and the highest performance of selecting a layered thin-film cannot be achieved based solely on a researcher's intuition. Human expert-based design is slow, and the performance of selecting an appropriate film is "un-perfect", especially in cases when the design target is complicated.

With the development of computer-aided design(CAD) technology, the inverse design has gained significant attention as a powerful approach to design layered optical thin-film without human experience. Several computational algorithms have been proposed to solve the inverse design problem, such as the evolutionary algorithm\cite{greiner1996robust},a genetic algorithm (GA)\cite{martin1995synthesis, li1997optical},the needle algorithm\cite{sullivan1996implementation, tikhonravov1994development}, and the particle swarm optimization (PSO)\cite{rabady2014global}. However, the design process based on computational 
algorithms is often time-consuming or computationally-intensive. For a complex nano-structure and board wavelength design target, these computational algorithms take a lot of computation time that is produced by the electromagnetic (EM) simulation, such as rigorous coupled-wave analysis (RCWA), the finite element method (FEM), the finite difference time domain (FDTD), and transfer matrix method(TMM). All of these methods are time and computationally expensive. In contrast, deep learning (DL) based algorithms, which are considered as being able to "learn" Maxwell's equations, were proposed to solve the nonlinear relationships between the structural parameters and film's performance by a large dataset\cite{peurifoy2018nanophotonic, liu2018training, asano2018optimization, malkiel2018plasmonic}. Learning the design process by human experts, and reinforcement learning are other solutions to solve this problem, which train an agent to learn about the parameter space of a series 
by exploration\cite{10.1016/j.solener.2019.12.013, undefined, jiang2020multilayer}. While the previously described algorithms have worked well for structural parameters optimization (nano-structure size and layer thickness) for a in thin-film inverse design well, there is little research on the design of component materials for these processes.

Generally, the simultaneous optimization of structural and material parameters is a combinatorial optimization problem. A challenging problem that arises in this field is material parameters, such as the complex refractive index $R(\lambda)=n+ik$, where $\lambda$ is the wavelength. The direct processing of high-dimensional features leads to additional time and poorer optimizer performance, or even absolute failure of the inverse design. One intuitive method is to code the materials, which is applied by inverse design methods on a very small scale\cite{10.1021/acsphotonics.7b01136, so2019simultaneous}. These two methods address the following materials' parameter optimization problem. The scope of these searches is complex and exponentially greater for a larger number of materials.
A variety of component materials, such as metals, semiconductors, alloys,  transparent material have been widely used in layered optical thin films to best match the design target. Simply encoding materials numerically in the form of key-value pairs, such as {(Au, 1), (Ag, 2), (Cu, 3)}, is not a scalable approach. For supervised learning-based inverse design methods, such massive class categories result in an insufficient number of samples that have been used in the training. For the reinforcement learning based inverse design methods, substantial materials result in large discrete action spaces, that bring reinforcement learning to a larger class of problems. Unsupervised deep neural networks, e.g. autoencoder(AE)\cite{rumelhart1985learning} and variation autoencoder(VAE)\cite{kingma2013auto}, work well for feature extraction. These methods achieve great success in generating abstract features with high dimensional data\cite{siradjuddin2019feature, kristjanpoller2014volatility}. Low-dimensional semantic space can be extracted by unsupervised-learning-based feature extraction methods from high-dimensional features by VAE-tSNE (variational autoencoder stochastic neighbor embedding) method\cite{Graving2020.07.17.207993}. In this technique, the model automatically learns a distribution of clusters and naturally creates a multi-scale representation. In this work, we use a VAE-tSNE feature reduction method to map the high dimensional complex refractive index of a material onto a two-dimensional semantic space. This method allows the environment space size of material not to increase with increasing of alternative materials.

We propose and test an implementation of an end-to-end reinforcement search system. This technique is the combination of an reinforcement learning and a genetic algorithm to design a layered optical thin-film from about 300 materials. The reinforcement learning environment space is based on a two-dimensional semantic space extracted by VAE-tSNE. In the following sections, we present the details of this algorithm’s formulation and search space. Furthermore, we use our proposed search system to design and optimize a layered solar absorber device.

\section{Design Target}
 The spectrum of an optical thin-film $S(\lambda|\mathbf{R}, \mathbf{d})=[Absorption(\lambda), Transmission(\lambda), Reflection(\lambda)]$ is determined by its component materials' refractive index $\mathbf{R}=[R_1,...,R_m]$, structural parameters(number of layers $m$, layer thickness $\mathbf{d}=[d_1, ...,d_m]$), and fabrication errors $\epsilon(\lambda)$. Fabrication errors often result from detects of the manufacturing equipment and errors in the manufacturing processes. In practice, as the wavelength range is continuous, the optimization design problem of thin-film can be described in Equation. (\ref{eq:1-1}), which is used to minimize the nonlinear least square problem between a given structure $S(\lambda | \mathbf{R}, \mathbf{d})$ and the target spectrum $S^*(\lambda)$.

\begin{equation}\label{eq:1-1}
    \underset{\boldsymbol{R},\boldsymbol{d}}{min} \sum_{\lambda} (\boldsymbol{S}(\lambda \vert \boldsymbol{R}, \boldsymbol{d}) - \boldsymbol{S^*}(\lambda))^2.
\end{equation}

\section{Method}
\begin{figure}[htbp]
\centering
\includegraphics[width=2.5in]{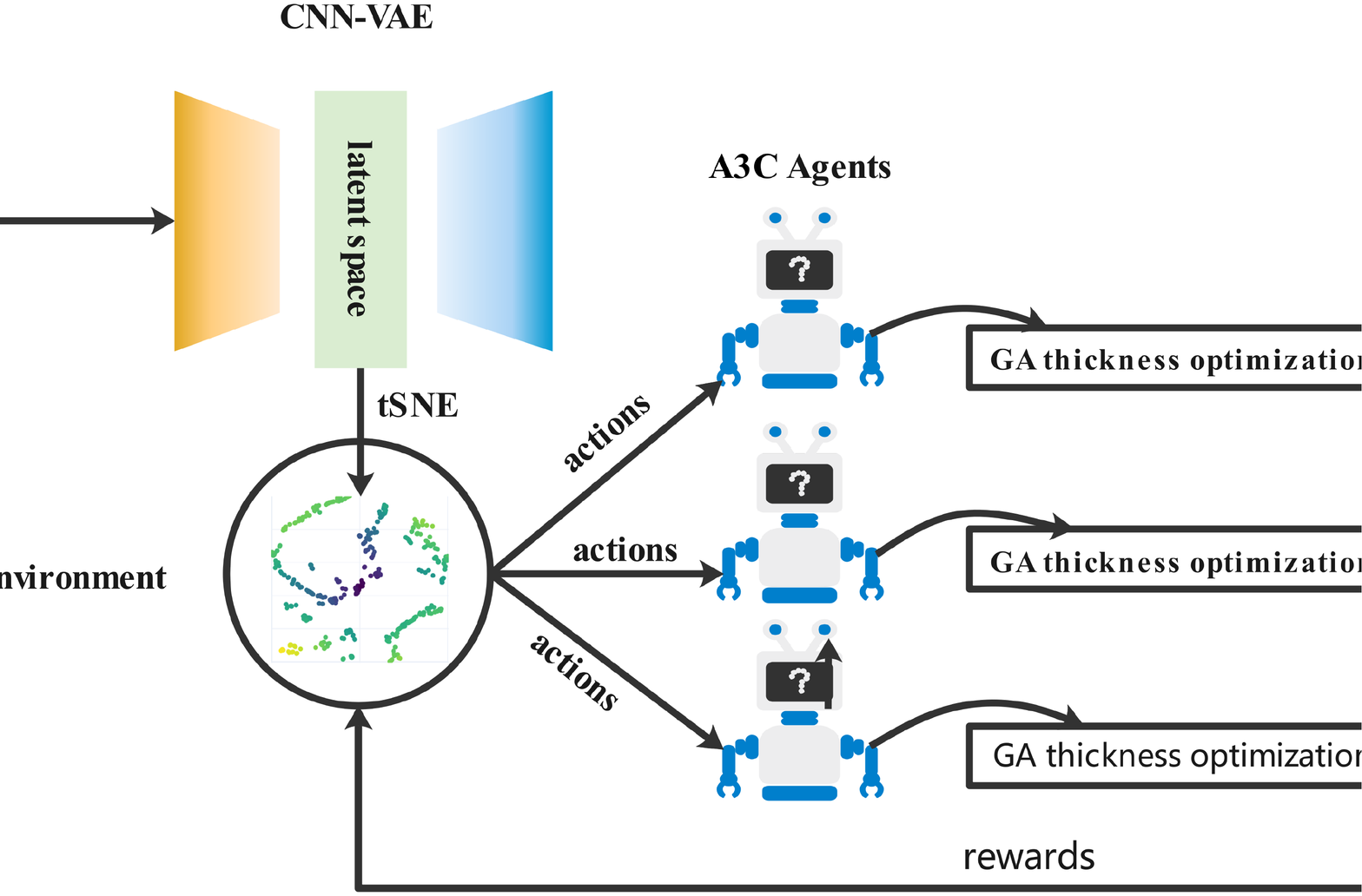}
\caption{The reinforcement search system for an optical thin-film inverse design. With the trained VAE-tSNE model, we mapped the optical constants of more than 300 optical materials into a 2-dimensional environment. A parallel reinforcement learning agents (A3C) adjusts the position of the material for different layers. In this system, GA is used to search the best thickness combinations from the results of each adjustment on the A3C agents. Each A3C's agent is a policy gradient agent to approximate the advantage function by neural networks. The searched film materials are given as feedback to the agents for further improvement of the performance.}\label{fig:sys_structure}
\end{figure}

Having defined the optimization problem in Eq.\ref{eq:1-1}, we can now outline the use of the memetic algorithm in finding the optimal multi-layer structure. Similar to the human design process, where well-designed films always improve on previous experience, the algorithmic optimization of optical films is performed step-by-step and can be considered as a Markov decision process (MDP). Reinforcement learning, a branch of machine learning, has been proposed to solve 
the MDP problem through an exploration-reward. In this paper, we use a reinforcement learning algorithm, called asynchronous advantage actor-critic (A3C), to find the best optical thin-film structure, This method allows running multiple agents in parallel instead of using only one while updating the shared network periodically and asynchronously. Fig.\ref{fig:sys_structure} shows the material selection system diagram.

%\begin{figure*}[!t]
%\centering
%\subfloat[Case I]{\includegraphics[width=2.5in]{box}%
%\hfil
%\subfloat[Case II]{\includegraphics[width=2.5in]{box}%
%\label{fig_second_case}}
%\caption{Simulation results for the network.}
%\label{fig_sim}
%\end{figure*}

\subsection{Environment Space}
\begin{figure}[htbp]
\centering
\includegraphics[width=4in]{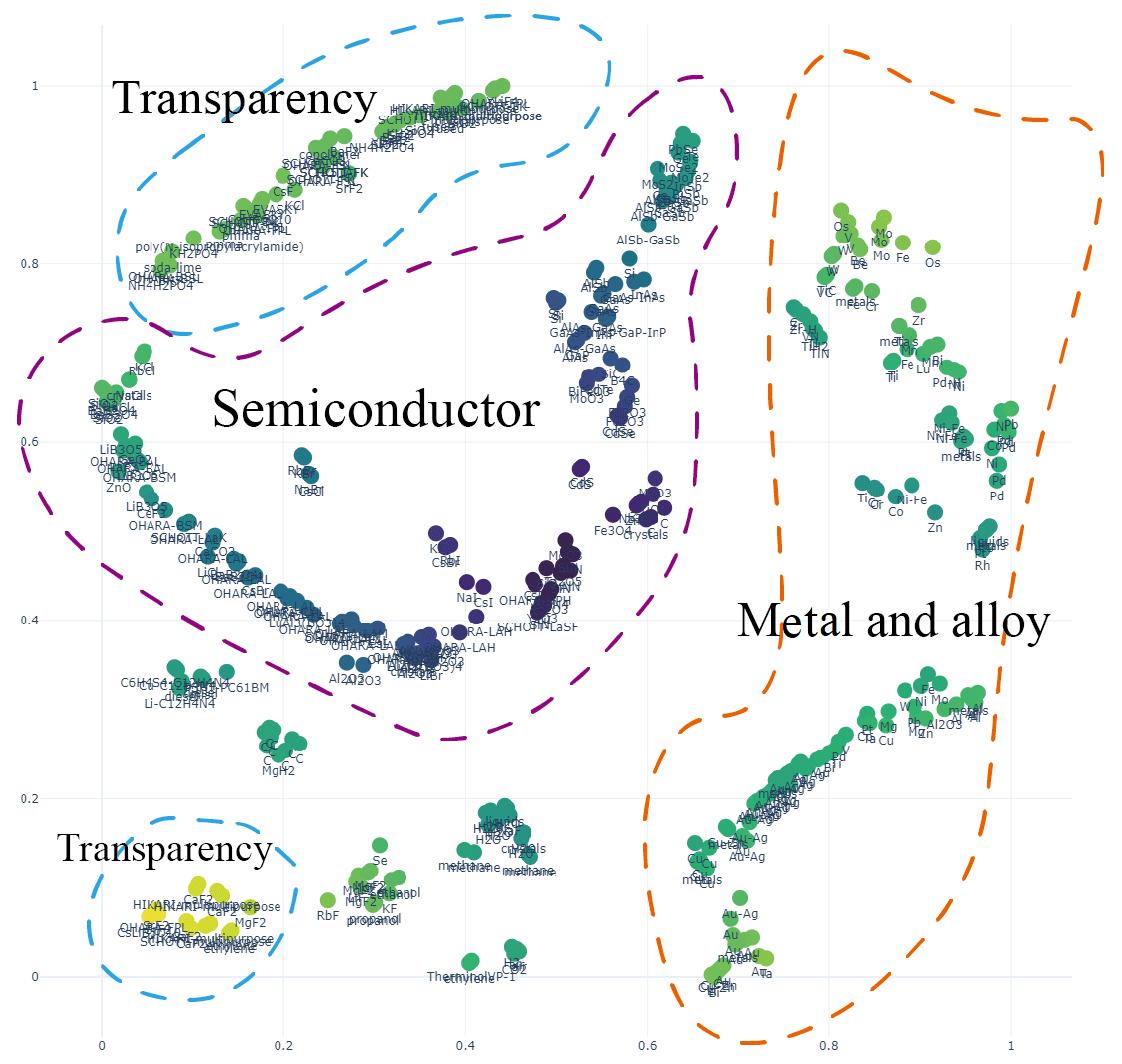}
\caption{Environment space generated from VAE-tSNE. Metal and alloys are distributed in the orange area. Dielectric semiconductors are distributed in the purple area. Transparency materials are distributed in the blue area.}\label{fig:mat_space}
\end{figure}

Making a learner-friendly environment is one of the central challenges faced by reinforcement learning to solve a specific problem. With a common reinforcement learning environment, as Go\cite{silver2007reinforcement}, Atari\cite{DBLP:journals/corr/BrockmanCPSSTZ16} and Duckietown\cite{gym_duckietown}, the agent explores and learns in a 2D or 3D environment space. The refractive index of the materials of an optical thin-film is continuous high-dimensional data, especially in the case of broadband optimization, and therefore cannot be directly characterized in 2D or 3D space. In this manuscript, we present a novel application of VAE-tSNE to embed the high-dimensional material refractive index to semantically relevant 2D latent variables. Fig. \ref{fig:mat_space} shows the environment space. Specifically, we trained 2-input 1D-CNN VAE with 5, 10, 15 and 20 units in a hidden layer. The latent space on the 20 units gets a minimum loss (0.003), after training 1,000 epochs. The tSNE embeds the 20 units latent vectors from the VAE generator for space building. By observing the environmental space, materials with similar properties are distributed in close proximity in this space after dimensionality reduction by VAE-tSNE.

\subsection{State}
The state is a 2D array of the material parameters for structure. The total number of possible states for a 5 layers film is, therefore, $100^2 * 100^2 * 100^2 * 100^2 * 100^2 = 10^{10}$. Manually searching all of these states is impossible. However, using A3C can produce desirable results in a reasonable time.

\subsection{Actions}
Actions determine the material changes that are needed to be applied to the optical thin-film. It is not feasible to set the position of material directly as action because the A3C agent is hard to train with a large number of discrete actions. To deal with this problem, the material actions are defined to change the material from the environment space. Table \ref{tab:actions} shows a list of all the actions for a design example of a 5-layer optical thin-film design.

\begin{table}[htbp]
\centering
\caption{Definition of actions used in A3C.}
\label{tab:actions}
\begin{tabular}{| c | c | c | c || c | c | c | c |}
\hline
Actions number & \# of layers & $\Delta x$ & $\Delta y$ & Actions number & \# of layers & $\Delta x$ & $\Delta y$ \\
\hline\hline
0  & 1 & 0.01 & 0 & 8  & 3 & -0.01 & 0 \\
\hline
1  & 1 & 0 & 0.01 & 9  & 3 & 0 & -0.01 \\
\hline
2  & 1 & -0.01 & 0 & 10  & 3 & 0.01 & 0 \\
\hline
3  & 1 & 0 & -0.01 & 11  & 3 & 0 & 0.01 \\
\hline
4  & 2 & 0.01 & 0 & 12  & 4 & -0.01 & 0 \\
\hline
5  & 2 & 0 & 0.01 & 13  & 4 & 0 & -0.01 \\
\hline
6  & 2 & -0.01 & 0 & 14  & 4 & 0 & -0.01 \\
\hline
7  & 2 & 0 & -0.01 & 15 & 4 & 0.01 & 0 \\
\hline

\end{tabular}
\end{table}

\subsection{Rewards}
Reward design enables the robustness of an RL system. To move our material design system in a certain and correct direction, we use both discrete-reward and continuous-reward. Considering that the environment space is sparse in the optimization process, we use this method of reward shaping to reduce the sparse payoff problem\cite{grzes2017reward}.

\begin{table}[htbp]
\centering
\caption{Definition of rewards used in A3C.}
\label{table:rewards_a3c}
\begin{tabular}{| c | c | c |}
\hline
Reward Number & Situation & Reward value \\
\hline\hline
1 & Film performance is not improved   & -1 \\ 
  & in the threshold step                &  \\ 
\hline
2 & Film performance is not improved & -0.01 \\
\hline
3 & Film performance is improved & Observation reward\\
\hline 
4 & Film performance meets the target & 1\\
\hline
\end{tabular}
\end{table}

Observation reward is an RMSE (root mean squarded error) loss type reward shaping function, which can be expressed as
\begin{equation}\label{eq:4-7}
    RMSE_{observation}(n,k,d)=\Sigma_{\lambda,\theta}W(\lambda)|S(\theta,\lambda;n,k,d) - S^*(\lambda,\theta)|.
\end{equation}

 Reward 0 and Reward 1 can restrict the agent from choosing meaningless pairwise actions (x minus 0.1, x plus 0.1) to gain a meaningless reward. The observation reward (Reward 3) provides the agents with a decaying reward system, which is designed to perform more exploitation at the beginning of each agent's action and more exploration at the end of each agent's action. 

\subsection{Proximity Material Matching}
\begin{figure}[htbp]
\centering
\includegraphics[width=0.3\textwidth]{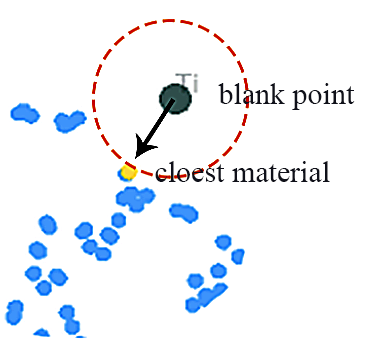}
\caption{Proximity material matching for Ti case. The agent walks to a blank point, the closet material is selected as the current material.}\label{fig:closet_mat}
\end{figure}

We can see from Fig. \ref{fig:mat_space} that our environment is sparse. This means that most of the environmental space does not correspond to a single material. To solve this problem, we define a rule that any point corresponding to a blank point is regarded as the closet material, as shown in Fig. \ref{fig:closet_mat}. This operation maps every point in the environment space to corresponding states. As a result the A3C agents will learn the policy of reinforcement learning.

\subsection{Thickness optimization}
In the thickness optimization process, to reduce the amount of computation in the overall material search, each material combination is optimized only once by the thickness optimization algorithm.  Compared with several traditional methods, as shown in Table \ref{tab:ga_comperation}, the genetic algorithm exhibited the best stability and best film performance in 50 attempts.

\begin{table}[htbp]
\centering
\caption{Results of the thickness optimization by traditional methods. Six heuristic optimization algorithms are used. The column "\emph{Best}" reports the largest value of absorption for each algorithm. \emph{Std} represents the stability of each algorithm. \emph{Time} is the average execution time to complete the optimization(in seconds).}
\label{tab:ga_comperation}

\begin{tabular}{p{1cm}p{1cm}p{1cm}p{0.8cm}p{0.8cm}p{0.8cm}p{0.8cm}p{0.8cm}p{0.8cm}p{0.8cm}p{0.8cm}p{0.8cm}}
\toprule
\multicolumn{3}{c}{} & \multicolumn{3}{c}{GA} & \multicolumn{3}{c}{GASA} & \multicolumn{3}{c}{PSO} \\
\cmidrule(lr){4-6} \cmidrule(lr){7-9} \cmidrule(lr){10-12} 

\multicolumn{3}{c}{Type} & Best & Std & Time & Best & Std & Time & Best & Std & Time \\   
\midrule
\multicolumn{3}{c}{8-layers solar absorber}  & 0.88 & 0.02 & 35 & 0.90 & 0.04 & 55 & 0.87 & 0.07 & 20\\

\midrule
\multicolumn{3}{c}{} & \multicolumn{3}{c}{SA} & \multicolumn{3}{c}{AFSA} & \multicolumn{3}{c}{DQN} \\
\cmidrule(lr){4-6} \cmidrule(lr){7-9} \cmidrule(lr){10-12}

\multicolumn{3}{c}{Type}  & Best & Std & Time & Best & Std & Time & Best & Std & Time \\   
\midrule
\multicolumn{3}{c}{8-layers solar absorber} & 0.87 & 0.04 & 210 & 0.81 & 0.02 & 376 & 0.94 & 0.01 & 1180\\

\bottomrule
\end{tabular}
\end{table}

\begin{figure}[htbp]
\centering
\includegraphics[width=0.6\textwidth]{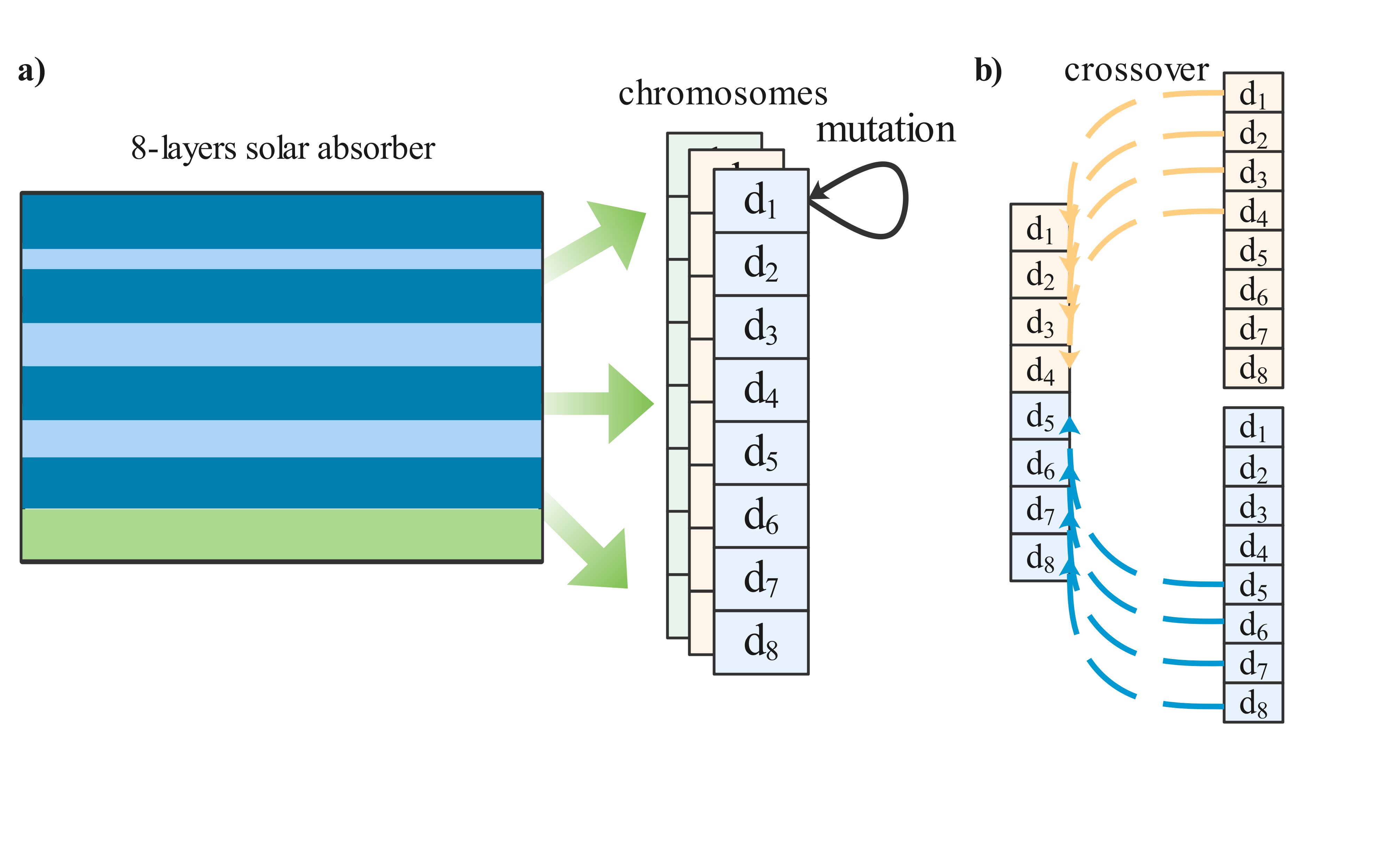}
\caption{Thickness optimization by GA. a) Encoding of thickness of optical thin-film and mutation operation. b) The crossover operation of encoded thickness of optical thin-film}
\label{fig:GA}
\end{figure}

In Fig. \ref{fig:GA} (a), we implement the genetic algorithm to optimize and improve the spectral performance of the layered optical thin-film with a population size of 100 and 500 generations. In the genetic thickness optimization algorithm, the chromosome represents the layer thickness of the thin-film. The variables of thickness are constructed from values that a boundary value ($10 nm-200 nm$) is assigned. Another four key parameters of GA are the selection rate (0.3), mutation rate (0.1), crossover rate (0.5) and the elitist selection rate (0.1), to improve the solution of the GA. For the crossover operation in Fig. \ref{fig:GA}  (b), a random number is generated to determine the crossover position. For the mutation operation, the thickness is a random number between the boundary value. The elitist selection operation selects the best 10\% chromosomes for the next generation in an iterative process, which is a slight variant of the general process of constructing a new population.

\subsection{A3C} 
A3C is a parallel policy gradient algorithm in reinforcement learning to learn a policy $\pi\left(a_{t}\mid{s}_{t}; \theta\right)$ and to estimate the value function $V\left(s_{t}; \theta_{v}\right)$ by agents. The policy and value function is updated by a mix of n-step returns(states and actions) when a terminal state is reached. The update that is performed by the algorithm can be seen as $\nabla_{\theta{'}}\log\pi\left(a_{t}\mid{s_{t}}; \theta{'}\right)A\left(s_{t}, a_{t}; \theta, \theta_{v}\right)$, where $A\left(s_{t}, a_{t}; \theta, \theta_{v}\right)$ is an estimate of the advantage function. The expression is given by:
$$
\sum^{k-1}_{i=0}\gamma^{i}r_{t+i} + \gamma^{k}V\left(s_{t+k}; \theta_{v}\right) - V\left(s_{t}; \theta_{v}\right).
$$
The value function is learned by the critics in A3C, while multi-actors are updated by the parameters in the master model. These updates are performed in parallel. After several episodes, the agents get synced with the master model and the parameters are updated. The gradients are accumulated as part of training for stability like the parallelized stochastic gradient descent in the deep learning training process. In this parallel architecture, each agent in the same episode is sample from different experiences. Such a mechanism can have a better chance of improving the optimization results for combinatorial optimization problems such as material selection problem.
\begin{figure}[htbp]
\centering
\includegraphics[width=0.6\textwidth]{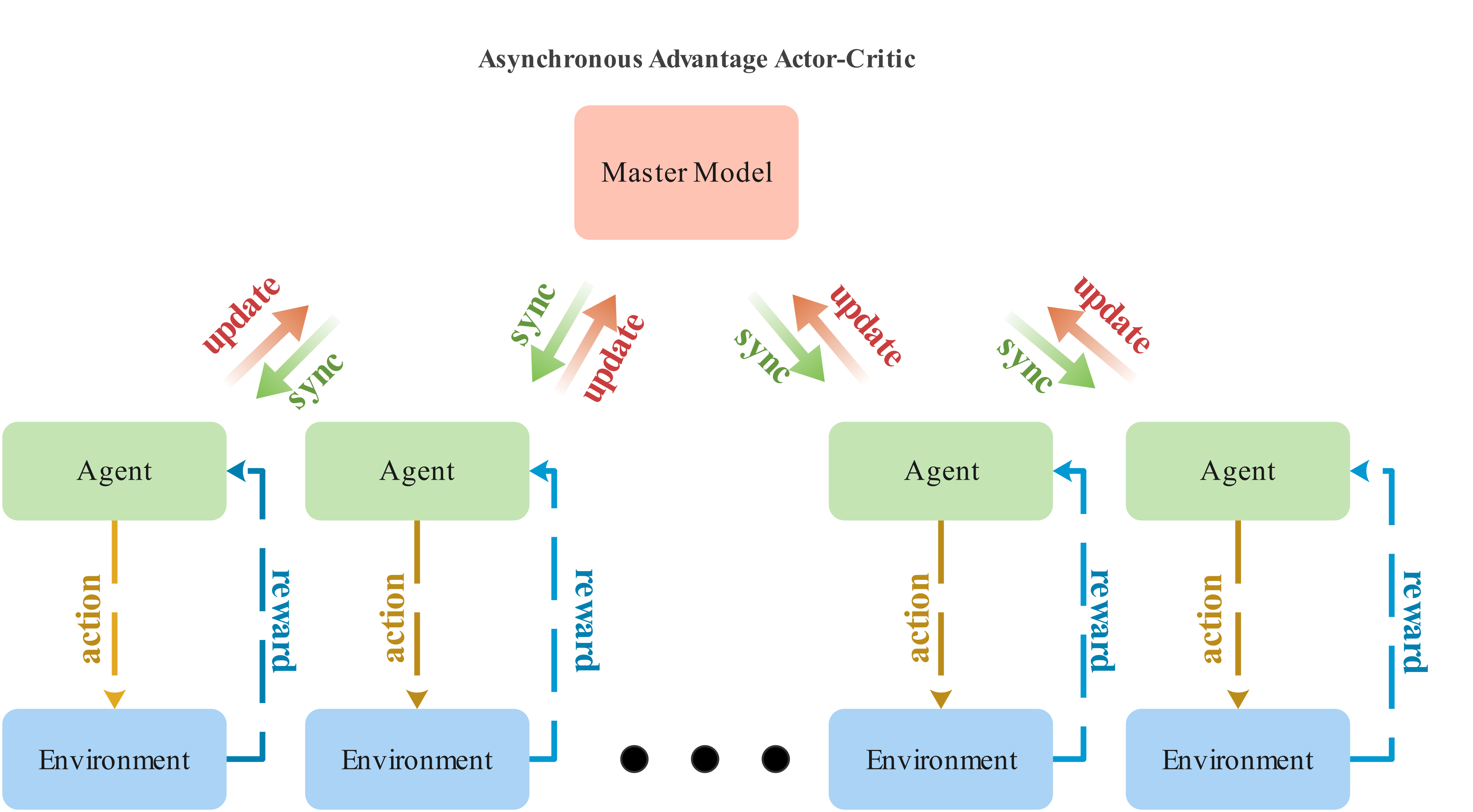}
\caption{The A3C algorithm begins by constructing the global network. This network will consist of layers to process spatial dependencies. Each agent have their own network and environment and run on a separate processor thread. The global network is constantly being updated by each of the agents, as they interact with their environment.}
\label{fig:A3C}
\end{figure}

In our implementation, the actor-network consists of 4 layers fully connected neural network with (5, 32, 16, 1) units. The critic-network consists of 5 layers fully connected neural network, with (5, 32, 16, 16 1) units. The activation function used in these two networks is the Relu. The Adam gradient descent for optimizing the networks.

\section{Results and Discussion}
\subsection{Optimization of Solar Absorber Device}
To demonstrate this material search system, we analyze a solar absorber device that can convert solar energy to thermal energy. This device uses multilayered thin-film structures consisting of alternating metal/alloys/semiconductor and dielectric layers.These layers have the advantage of excellent spectral properties in both broad solar spectral and wide incident angle regions, low thermal emittance, and high thermal stability. 

\begin{figure}[!t]
\centering
\includegraphics[width=0.8\textwidth]{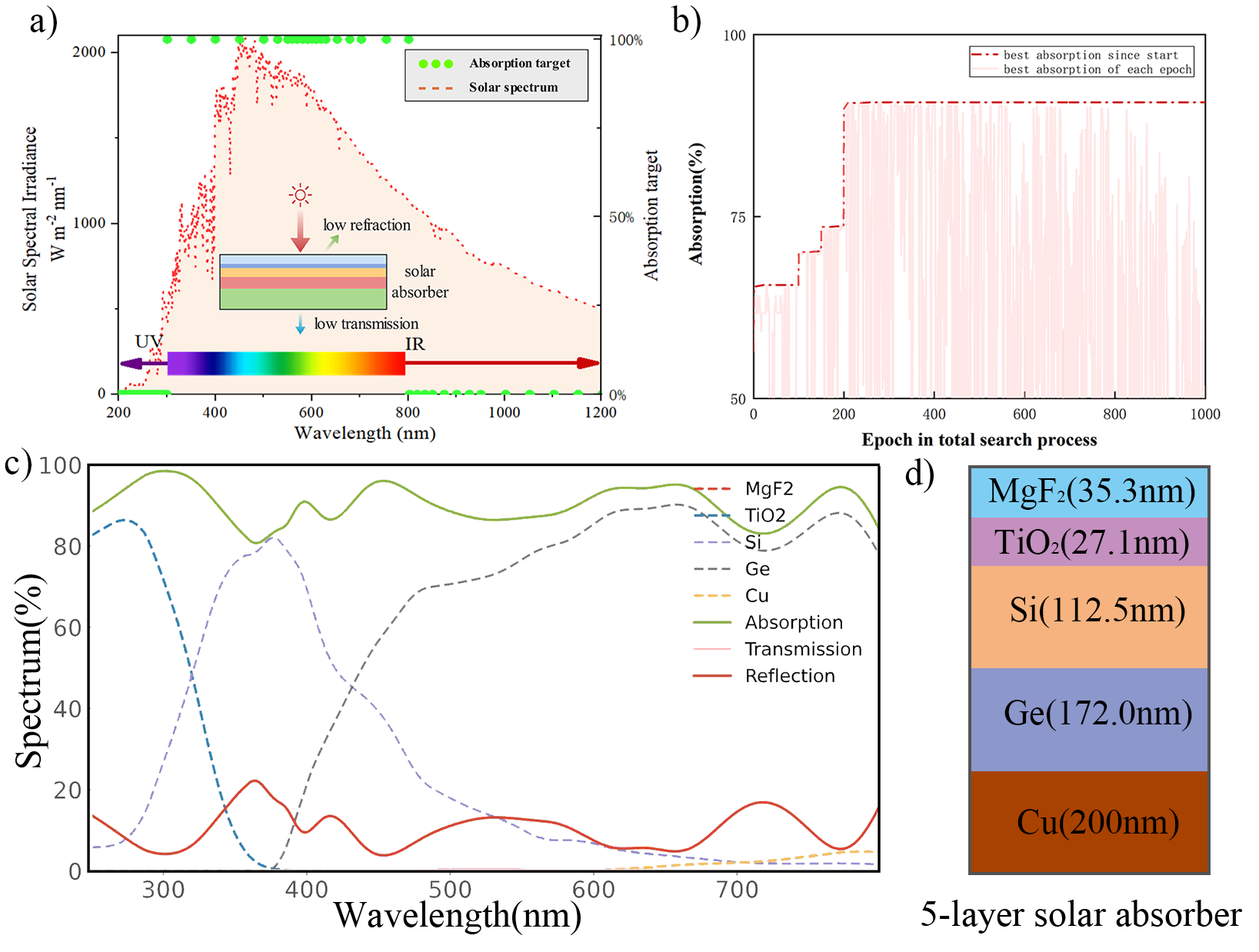}
\caption{The design of solar absorber by our proposed search system. a) The target absorptivity spectrum of the solar absorber is compared to the solar spectrum. The operating principles of the multi-layer solar selective absorber. b) The best absorption in the search process. c) The absorption spectrum of designed film d) The material composition and thicknesses for the material search system}
\label{fig:4in1}
\end{figure}

 We have performed a complete study of 4/6/8-layer SSR thin-film devices in our previous research.\cite{hu2018multilayered, Li:07, Zhou:12, Hu_2017}. A SSR with 5-layer structures have not been extensively studied.  The SSR with 5-layer structures have not been extensively studied. Fig.\ref{fig:4in1}(a)  shows that solar energy is highest near the visible wavelengths. At an angle of normal light, the goal is 100\% absorption at the wavelength range from 250 nm to 800 nm and no absorption at other wavelengths. We aim to achieve this performance with our studies.

From our search for 1000 epochs, we plot in Fig. \ref{fig:4in1}(b) the best and cumulative absorption in each epoch. The agents develop the best material structure for a 5-layer structure with the material composition the thickness shown in Fig. \ref{fig:4in1}(d). The algorithm selects materials in the following order $[MgF_2, TiO_2, Si, Ge, Cu]$,respectively, thickness of each layer with the following order $[35.3nm, 27.1nm, 112.5nm, 172.0nm, 200.0nm]$. The designed absorber can achieve broadband absorption due to the characteristics of selected materials as shown in Fig. \ref{fig:4in1}(c).  The average absorption in the wavelength range, 250 nm to 800 nm, is above 91\%. In contrast to other thin film optimization algorithms that can only handle thickness optimization problems, our algorithm selects effective constituent materials from a huge range of materials.

% \begin{center}
% \begin{table}[htbp]
% \centering
% \caption{Material composition and thicknesses for the material search system.}
% \label{table:Optimal Target}
% \begin{tabular}{| c | c | c |}
% \hline
% Layer # & Material & Thickness(nm) \\
% \hline
% \hline
%     &  Air  & Topstrate  \\
% \hline
% 1   &   $MgF_2$    &    35.3        \\
% \hline
% 2   &   $TiO_2$    &    27.1       \\
% \hline
% 3   &   $Si$    &   112.5        \\
% \hline
% 4   &   $Ge$    &   172.0        \\
% \hline
% 5   &   $Cu$    &   200          \\
% \hline 
%     &   $Glass$ &  Substrate     \\
% \hline
% \end{tabular}
% \end{table}
% \end{center}

\section{Conclusion}
To conclude, this paper has created an RL based on A3C and GA to optimize the design of a multi-layer solar absorber. Using the VAE-tSNE method with more than 800 different materials, this technique selected the most appropriate one. The material search system can automatically search material composition and structure to achieve a target optical spectrum. Compared with previous work, our approach can handle and select material from a significantly larger dataset without any human intervention. As a demonstration, we have used it to design a solar selective absorber. By using a diverse set of materials, the multi-layer structure has excellent solar absorption of 91\% in visible light. Because of its versatility and effectiveness, our material search system proves to be a state of the art tool for multi-layer optical thin-film inverse design.  It can be not only used for thermal and energy applications, but also for other optical device designs and optimization.

\bibliographystyle{IEEEtran}
\bibliography{paper}
\end{document}